\documentclass[runningheads]{llncs}

\usepackage{eccv}

\usepackage{eccvabbrv}

\usepackage{float}
\usepackage{graphicx}
\usepackage{booktabs}
\usepackage{multirow}
\usepackage{amsmath}
\usepackage{xcolor} 
\usepackage{pifont} 
\usepackage{color, colortbl} 
\usepackage{arydshln} 
\usepackage{tabularx} 
\usepackage{footnote} 

\usepackage[accsupp]{axessibility}  

\usepackage[pagebackref,breaklinks,colorlinks,citecolor=eccvblue]{hyperref}

\usepackage{orcidlink}

\begin{document}

\title{\includegraphics[width=0.05\textwidth]{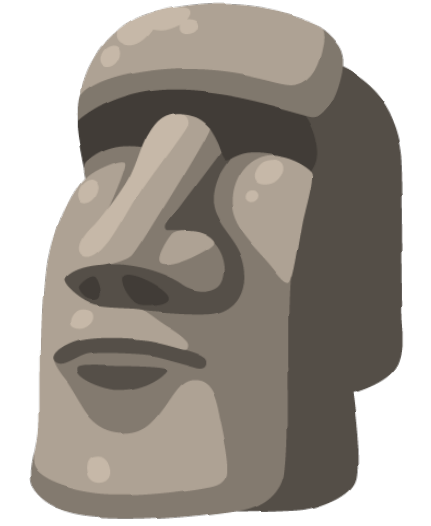} MoAI: Mixture of All Intelligence\\for Large Language and Vision Models} 

\titlerunning{MoAI: Mixture of All Intelligence for Large Language and Vision Model}

\author{Byung-Kwan Lee \and
Beomchan Park \and
Chae Won Kim \and
Yong Man Ro}

\authorrunning{B.K. Lee, B. Park, C.W. Kim, and Y.M. Ro}

\institute{School of Electrical Engineering\\
Korea Advanced Institute of Science and Technology (KAIST)\\
\email{\{leebk, bpark0810, chaewonkim, ymro\}@kaist.ac.kr}}

\maketitle

\begin{abstract}
  The rise of large language models (LLMs) and instruction tuning has led to the current trend of instruction-tuned large language and vision models (LLVMs). This trend involves either meticulously curating numerous instruction tuning datasets tailored to specific objectives or enlarging LLVMs to manage vast amounts of vision language (VL) data. However, current LLVMs have disregarded the detailed and comprehensive real-world scene understanding available from specialized computer vision (CV) models in visual perception tasks such as segmentation, detection, scene graph generation (SGG), and optical character recognition (OCR). Instead, the existing LLVMs rely mainly on the large capacity and emergent capabilities of their LLM backbones. Therefore, we present a new LLVM, \textbf{M}ixture \textbf{o}f \textbf{A}ll \textbf{I}ntelligence (\includegraphics[width=0.025\textwidth]{figure/moai.pdf} \textbf{MoAI}), which leverages auxiliary visual information obtained from the outputs of external segmentation, detection, SGG, and OCR models. MoAI operates through two newly introduced modules: \textit{MoAI-Compressor} and \textit{MoAI-Mixer}. After verbalizing the outputs of the external CV models, the MoAI-Compressor aligns and condenses them to efficiently use relevant auxiliary visual information for VL tasks. MoAI-Mixer then blends three types of intelligence—(1) visual features, (2) auxiliary features from the external CV models, and (3) language features—utilizing the concept of Mixture of Experts. Through this integration, MoAI significantly outperforms both open-source and closed-source LLVMs in numerous zero-shot VL tasks, particularly those related to real-world scene understanding such as object existence, positions, relations, and OCR without enlarging the model size or curating extra visual instruction tuning datasets. Code is available in \href{https://github.com/ByungKwanLee/MoAI}{https://github.com/ByungKwanLee/MoAI}.
\end{abstract}

\section{Introduction}
\label{sec:intro}
Combining large language models (LLMs) such as PaLM~\cite{chowdhery2023palm} and T5~\cite{raffel2020exploring} with instruction tuning datasets from Flan~\cite{wei2022finetuned}, Chung \textit{et al.}~\cite{chung2022scaling} has developed Flan-PaLM and Flan-T5 for instruction-tuned LLMs. These models leverage an expanded instruction tuning dataset covering various tasks, and have been further scaled up to enlarge their capacities, resulting in notable improvements in zero-shot performance across numerous language tasks.

Alongside the success of the instruction-tuned LLMs, several visual instruction tuning datasets~\cite{liu2023visual, dai2023instructblip, chen2023sharegpt4v, bai2023qwen, wang2023cogvlm} have been meticulously curated to enhance zero-shot vision language (VL) performances in large language and vision models (LLVMs). Furthermore, concerted efforts have been made to substantially scale up LLVMs~\cite{wang2023cogvlm, bai2023qwen, Yi, liu2024llavanext}, aiming for strong zero-shot performances in VL datasets. With the extension of visual instruction tuning datasets and the scaling up of LLVMs, open-source LLVMs~\cite{liu2023visual, dai2023instructblip, chen2023sharegpt4v, bai2023qwen, wang2023cogvlm, Yi, liu2024llavanext, chen2023minigpt, zhu2023minigpt, gong2023multimodal, zhang2023internlm} have been closing the gap in zero-shot VL performances compared to closed-source LLVMs such as GPT-4V~\cite{gptsyscard, gpttechnical}, Gemini-Pro~\cite{team2023gemini}, and Qwen-VL-Plus~\cite{bai2023qwen}.

\begin{figure}[t!]
\vspace{-3mm}
    \centering
    \includegraphics[width=0.95\textwidth]{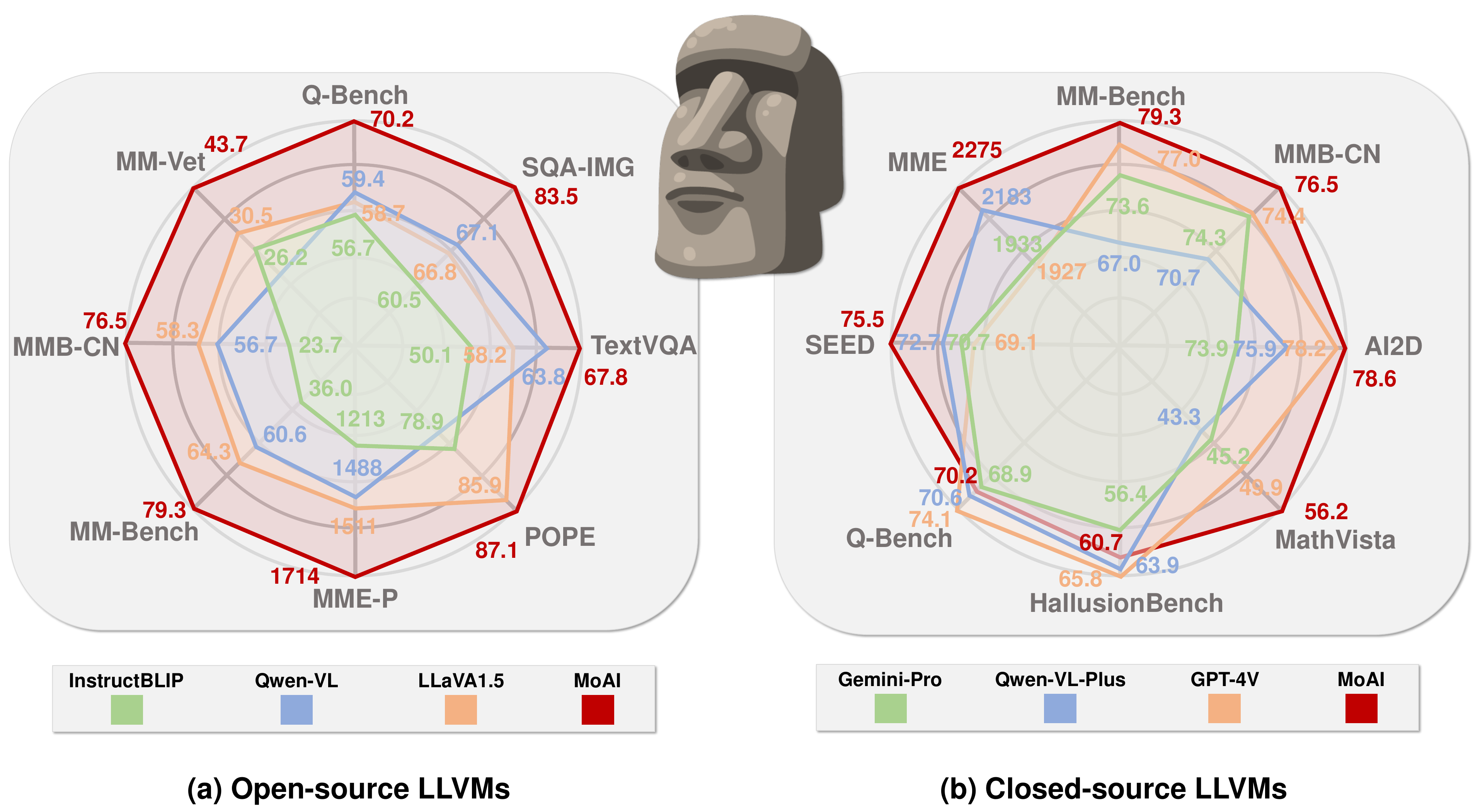}
    \vspace{-3mm}
    \caption{Comparing the scores and accuracies of numerous VL benchmarks for various open-source and closed-source LLVMs with those for \includegraphics[width=0.025\textwidth]{figure/moai.pdf} \textbf{MoAI}.}
    \label{fig:llvms_open_close}
    \vspace{-5mm}
\end{figure}

However, current open-source LLVMs have not explicitly or fully leveraged detailed and comprehensive real-world scene understanding, relying mainly on the large capacity and emergent capabilities of their LLM backbones. Several studies in cognitive science and machine learning~\cite{BIEDERMAN1982143, NEURIPS2023_efe406d6, epstein2019scene} argue that fundamental scene perception ability may stem from various cognitive functions, including recognizing object presence, determining their positions, identifying their states, understanding their relationships, extracting spatial scene layouts, and grasping non-object notions which may include written texts. Fortunately, these cognitive functions can be acquired from specialized computer vision (CV) models which have been researched and developed over decades for visual perception tasks such as segmentation~\cite{cheng2022masked, jain2023oneformer}, detection~\cite{zong2023detrs, minderer2023scaling}, scene graph generation (SGG)~\cite{yang2022panoptic, kim2024adaptive}, and optical character recognition (OCR)~\cite{du2021pp, li2023trocr}.

Shifting the focus from instruction-tuning to utilizing these external CV models is expected to enhance the real-world scene understanding of LLVMs, covering object existence, positions, relations, and OCR. Recognition of objects and their positions~\cite{lee2024collavo} can be facilitated by panoptic segmentation and open-world object detection models. For a more comprehensive understanding, involving object states and relationships (\textit{i.e.,} compositional reasoning~\cite{NEURIPS2023_efe406d6}), a scene graph generation (SGG) model is necessary. Moreover, text descriptions within an image as a non-object notion can be recognized through an OCR model.

\begin{figure}[t]
\vspace{-3mm}
    \centering
    \includegraphics[width=0.95\textwidth]{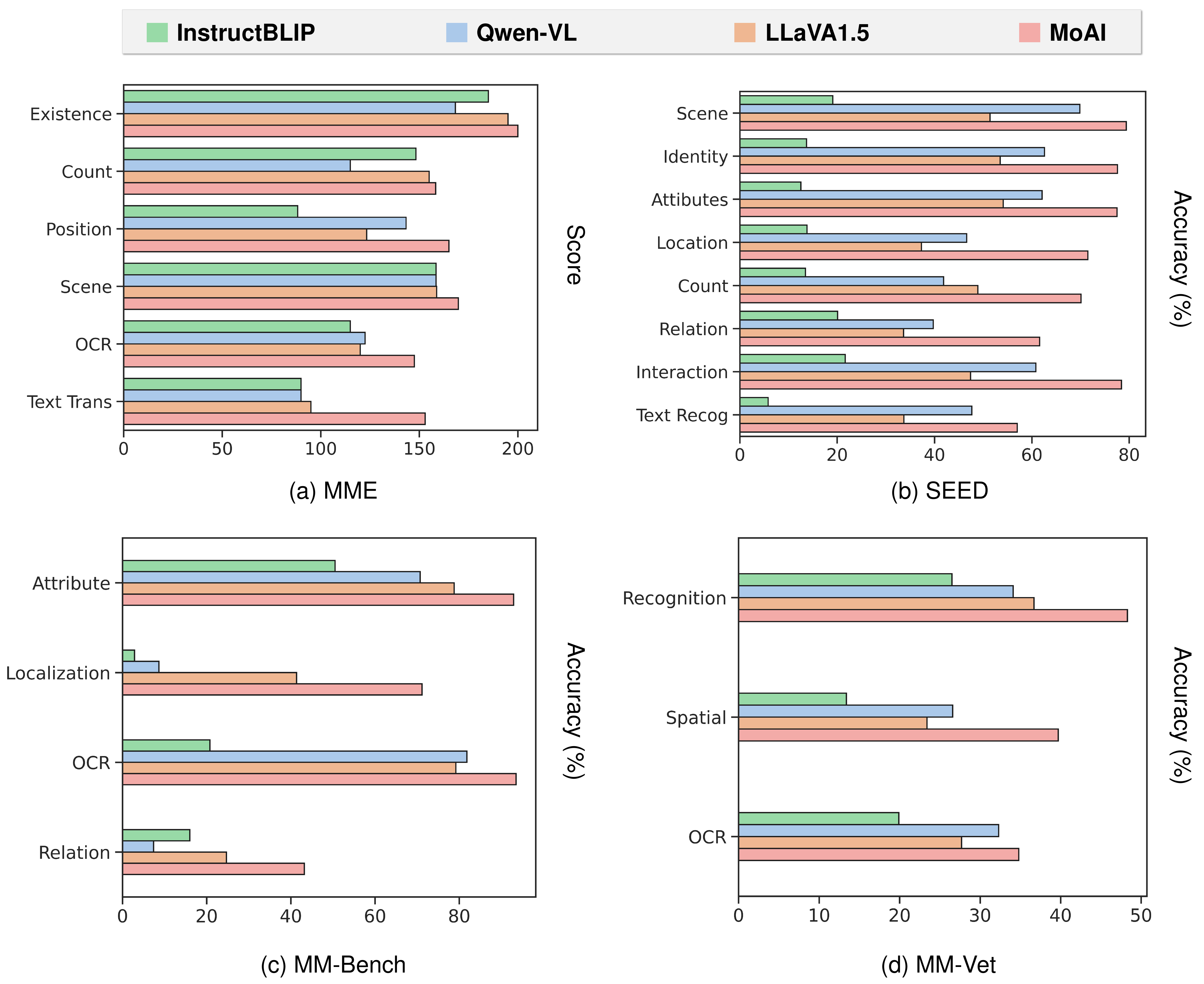}
    \vspace{-3mm}
    \caption{Comparing the scores and accuracies of dimensions related to real-world scene understanding in MME~\cite{fu2023mme}, SEED~\cite{li2023seed}, MM-Bench~\cite{liu2023mmbench}, and MM-Vet~\cite{yu2023mm} for validating capabilities of various LLVMs such as InstructBLIP~\cite{dai2023instructblip}, Qwen-VL~\cite{bai2023qwen}, and LLaVA1.5~\cite{liu2023improved}.}
    \label{fig:perception}
    \vspace{-5mm}
\end{figure}

In light of this, we propose a new LLVM, \textbf{M}ixture \textbf{o}f \textbf{A}ll \textbf{I}ntelligence (\includegraphics[width=0.025\textwidth]{figure/moai.pdf} \textbf{MoAI}), which leverages auxiliary visual information obtained from various sources: (1) panoptic segmentation~\cite{cheng2022masked}, (2) open-world object detection~\cite{minderer2023scaling}, (3) SGG~\cite{yang2022panoptic}, and (4) OCR~\cite{du2021pp} models. To effectively leverage this information, we introduce two new modules: \textit{MoAI-Compressor} and \textit{MoAI-Mixer}. The MoAI-Compressor aligns and condenses the verbalized outputs of the external CV models into auxiliary visual information, enabling the efficient use of relevant information for VL tasks. Subsequently, \textit{MoAI-Mixer} blends three types of intelligence—(1) visual features, (2) auxiliary features from external CV models, and (3) language features—into a cohesive whole.

In constructing the MoAI-Mixer, we draw inspiration from the concept of Mixture of Experts (MoE)~\cite{shazeer2017, riquelme2021scaling, zhou2022mixture, mustafa2022multimodal}. Our challenge lies in seamlessly integrating original features (\textit{i.e.,} visual and language features) used in the multimodal language model (MLM) of MoAI—an LLM backbone that takes visual tokens outputted by the visual encoder along with text tokens—with auxiliary features acquired from external CV models and the MoAI-Compressor. We employ cross- and self-attention modules to construct six expert modules in the MoAI-Mixer, covering the three types of aforementioned intelligence. Furthermore, we utilize gating networks to determine the optimal combination of weights for these expert modules.

By combining the MoAI-Compressor and MoAI-Mixer, MoAI effectively utilizes outputs from external CV models and mix three sources of intelligence, thereby enhancing its visual perception capabilities for tackling complex question answering tasks. As depicted in \cref{fig:perception}, our results demonstrate that MoAI has significantly outperformed in visual perception scores three strong LLVM baselines: InstructBLIP~\cite{dai2023instructblip}, Qwen-VL~\cite{bai2023qwen}, LLaVA1.5~\cite{liu2023improved}, even without additional curation of visual instruction tuning datasets or scaling up LLVMs. Furthermore, owing to its improved visual perception ability, MoAI exhibits potent zero-shot performances in VL tasks, surpassing closed-source LLVMs, as illustrated in \cref{fig:llvms_open_close}. The success of MoAI is attributed to its utilization of diverse auxiliary visual information from external CV models and the integration of three intelligence types to effectively execute VL tasks. Our contribution can be summarized in two main aspects as follows:

\begin{itemize}
    \item We introduce a new large language and vision model, \includegraphics[width=0.025\textwidth]{figure/moai.pdf} \textbf{MoAI}, which handles various auxiliary visual information from external CV models (\textit{MoAI-Compressor}) and blends three types of intelligence (\textit{MoAI-Mixer}). 
    \item \includegraphics[width=0.025\textwidth]{figure/moai.pdf} \textbf{MoAI} stands out for its exceptional visual perception ability in VL tasks, surpassing both open-source and closed-source LLVMs in zero-shot VL performances. This ability is achieved by considering detailed and comprehensive real-world scene understanding without requiring scaling up either the model size or dataset size.
\end{itemize}

\section{Related Works}
\label{sec:related works}

\paragraph{\textbf{LLMs and LLVMs.}}
LLMs have emerged alongside their competent generalization capability and the effectiveness of instruction tuning datasets. GPTs~\cite{radford2018improving,radford2019language,brown2020language} played a crucial role in paving the way for LLMs by demonstrating strong zero-shot or few-shot performance across various language tasks, including text classification, question answering, machine translation, complex reasoning tasks, and so on. These generalization abilities of LLMs have been achieved by enormously increasing both model capacities and training datasets, as seen in works such as T5~\cite{raffel2020exploring}, PaLM~\cite{chowdhery2023palm}, OPT~\cite{zhang2022opt}. The progress in training methods and datasets further enhances the zero-shot generalization of LLMs, transitioning from large-scale pre-training datasets to instruction tuning datasets~\cite{wei2022finetuned, chung2022scaling, ouyang2022training, iyer2022opt}. Instruction tuning~\cite{wei2022finetuned} enables LLMs to follow instructions in human natural language under complex real-world scenarios. Instruction-tuned LLMs, such as Flan-T5, Flan-PaLM~\cite{chung2022scaling}, OPT-IML~\cite{iyer2022opt}, and InstructGPT~\cite{ouyang2022training}, clearly demonstrate the effectiveness of instruction tuning. Researchers have taken a step further by applying similar strategies to multimodal counterparts, LLVMs, which consist of a visual encoder and a backbone multimodal language model (MLM). For example, LLaVA~\cite{liu2023visual} and ShareGPT4V~\cite{chen2023sharegpt4v} utilize GPT-4~\cite{achiam2023gpt} and GPT-4V~\cite{gptsyscard,gpttechnical}, respectively, to create visual instruction tuning datasets, while others~\cite{dai2023instructblip, bai2023qwen, wang2023cogvlm} have also developed various visual instruction tuning datasets for their own unique objectives. However, the existing LLVMs have overlooked the detailed and comprehensive real-world scene understanding available from CV models with great advancements over the last decades. The CV models have been overshadowed by LLVMs with enlarged capacities and visual instruction tuning datasets in the era of LLVMs. From this perspective, MoAI highlights the effectiveness of utilizing auxiliary visual information obtained from external CV models, showing enhanced visual perception capabilities for VL benchmarks.

\paragraph{\textbf{Mixture of Experts.}}
Jacobs \textit{et al.}~\cite{jacobs1991adaptive} has first introduced the concept of Mixture of Experts (MoE) to machine learning, where separate networks called `experts' handle different segments of the input space, and each segment is guided to relevant experts by a gating network. This idea is further developed by deep MoE~\cite{eigen2013learning} where MoE layers are stacked in depth, and by conditional computation~\cite{bengio2013estimating} where only a few experts are conditionally activated by a given input. In modern deep learning, Shazeer \textit{et al.}~\cite{shazeer2017} integrates an MoE layer with LSTMs~\cite{hochreiter1997long} where a gating network independently routes each token to selectively activated experts.  This integration enhances performance in language modeling and machine translation tasks. Furthermore, Switch Transformers~\cite{fedus2022switch} merge an MoE layer and Transformers~\cite{vaswani2017attention} by replacing a dense feed forward network (FFN) inside a Transformer layer with multiple experts and a gating network, paving a way to the successful use of MoE in Transformer-based LLVMs such as MoE-LLaVA~\cite{lin2024moe}. The philosophy of MoE in deep learning is to enlarge model capacity without sacrificing computational efficiency~\cite{eigen2013learning, shazeer2017, fedus2022switch, zoph2022st, komatsuzaki2022sparse, lin2024moe, jiang2024mixtral}. On the other hand, we focus on a different yet fundamental aspect of MoE, where we intend that each expert is designed to specialize in a particular segment of input. While previous MoE methods do not explicitly assign roles to individual experts and instead expect specialization to emerge during optimization, MoAI designates cross- and self-attention modules as experts and learns them explicitly to mix information across modalities (\textit{i.e.,} visual, auxiliary, and language features). Specifically, MoAI facilitates pairs of (1) visual-auxiliary feature, (2) visual-language feature, (3) visual-visual feature, (4) language-auxiliary feature, (5) language-visual feature, and (6) language-language feature. Each pair is considered as a query-key pair for a respective cross- or self-attention module serving as experts, clarifying the fusion of information across diverse modalities.

\section{\includegraphics[width=0.04\textwidth]{figure/moai.pdf} MoAI: Mixture of All Intelligence}
\label{sec:moai}

\begin{figure}[t]
\vspace{-3mm}
    \centering
    \includegraphics[width=0.95\textwidth]{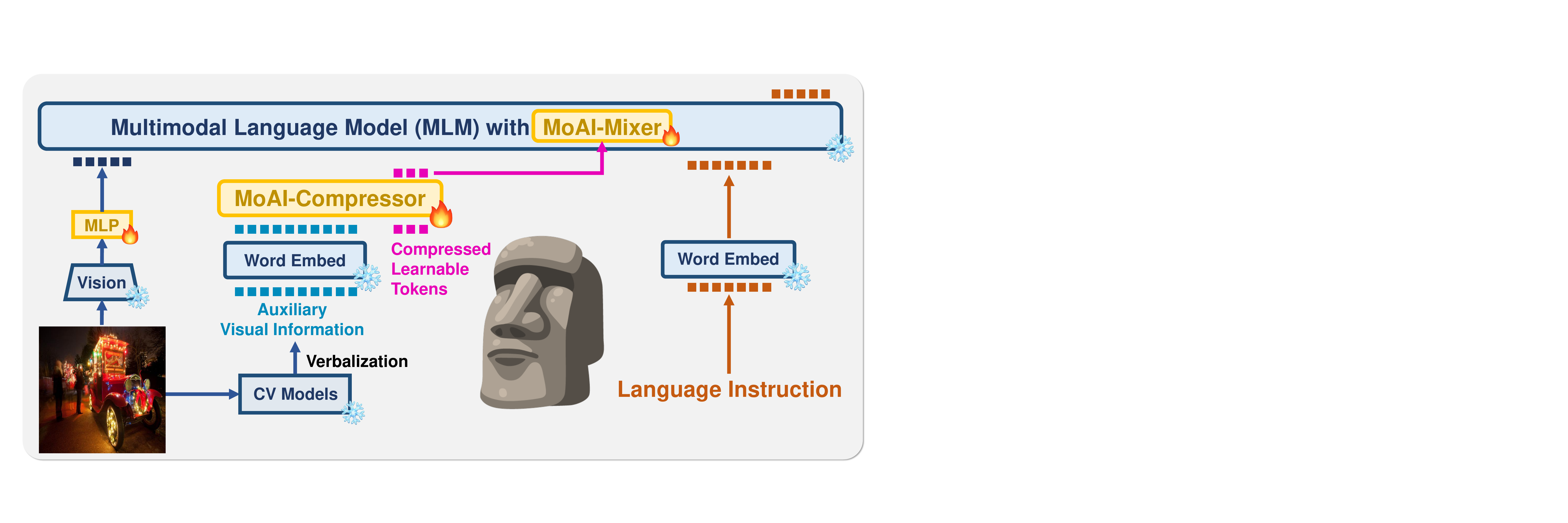}
    \vspace{-3mm}
    \caption{Overview of \includegraphics[width=0.025\textwidth]{figure/moai.pdf} \textbf{MoAI} architecture. Compressed learnable tokens, the parameters of MoAI-Compressor and MoAI-Mixer are learned. `Vision' represents vision encoder to embed visual features and ice/fire symbols represent the modules to freeze or learn. Note that, `Word Embed' represents the word embedding dictionary of MLM.}
    \label{fig:moai_arch}
    \vspace{-5mm}
\end{figure}

\begin{figure}[t!]
\vspace{-3mm}
    \centering
    \includegraphics[width=0.9\textwidth]{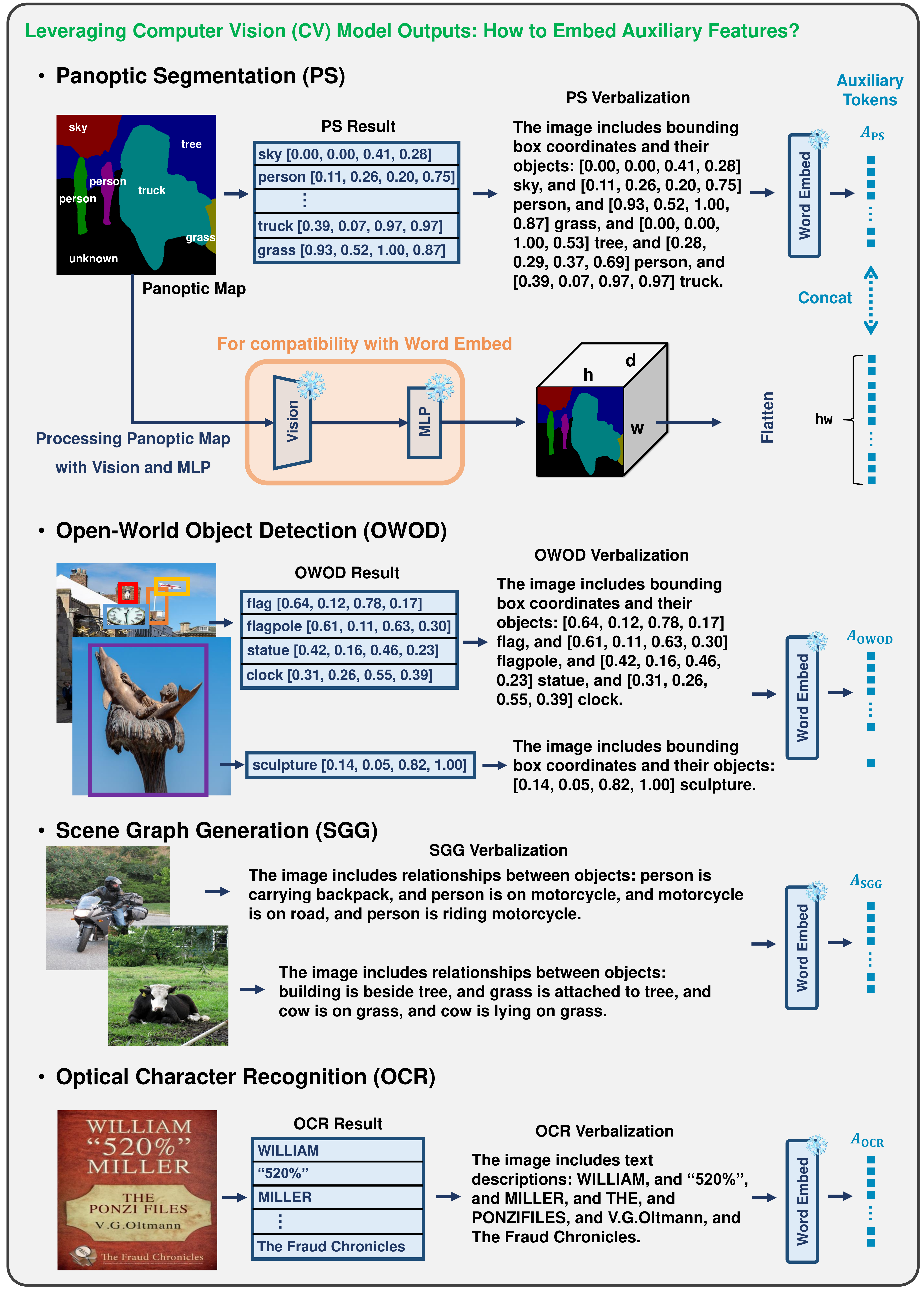}
    \vspace{-3mm}
    \caption{Verbalization process of \includegraphics[width=0.025\textwidth]{figure/moai.pdf} \textbf{MoAI} for external CV models: panoptic segmentation (PS), open-world object detection (OWOD), scene graph generation (SGG), and optical character recognition (OCR). Note that, `d' denotes channel dimension of MLM, thus auxiliary tokens have equal channel dimension.}
    \label{fig:verbalization}
    \vspace{-5mm}
\end{figure}

\paragraph{\textbf{Model Architecture.}} As depicted in \cref{fig:moai_arch}, MoAI consists of a vision encoder, a backbone multimodal language model (MLM) equipped with MoAI-Mixers, intermediate MLP connectors between the vision encoder and MLM, and a MoAI-Compressor which leverages four external computer vision (CV) models for panoptic segmentation~\cite{cheng2022masked}, open-world object detection~\cite{minderer2023scaling}, scene graph generation (SGG)~\cite{yang2022panoptic}, and optical character recognition (OCR)~\cite{du2021pp}. MoAI-Compressor is introduced to process diverse auxiliary visual information acquired from the external CV models, where the CV model outputs are processed via verbalization as shown in \cref{fig:verbalization} to make them aligned and interpretable to the MLM utilized in MoAI. In addition, MoAI-Mixer is further presented to efficiently harmonize original two features (\textit{i.e.,} visual and language features) with auxiliary features from the external CV models. The details of verbalization, MoAI-Compressor, and MoAI-Mixer will be explained in this section.

\paragraph{\textbf{Vision and Language Backbone.}} CLIP-L/14~\cite{clip} is selected as the vision encoder, due to its guaranteed proficiency in image understanding aligned with text for vision language  tasks~\cite{liu2023visual, liu2023improved, chen2023sharegpt4v, liu2024llavanext}. The MLM utilized in MoAI is based on InternLM2-7B~\cite{cai2024internlm2}, which is a multilingual foundation model instruction-tuned by multilingual datasets with 1.6T tokens through a series of progressive pretraining phases and reinforcement learning from human feedback (RLHF)~\cite{christiano2017deep, stiennon2020learning, ouyang2022training}. Two linear layers with GELU activation function~\cite{hendrycks2016gaussian} serve as the bridge connector between vision and language components, denoted by `MLP' in \cref{fig:moai_arch}.

\paragraph{\textbf{Verbalization.}} Since a multimodal language model (MLM) is adopted to construct MoAI, we convert CV model outputs into natural language format in order to make them understandable to the MLM through a process called verbalization. \cref{fig:verbalization} illustrates how the four CV model outputs undergo verbalization alongside the creation of auxiliary tokens semantically aligned to the MLM.

A panoptic segmentation model enables us to distinguish foreground and background objects in an image at once. Furthermore, we can compute bounding box coordinates (\textit{e.g.,} $[x_{\text{min}}, y_{\text{min}}, x_{\text{max}}, y_{\text{max}}]$) from the segmentation map. Consequently, verbalizing the outputs from panoptic segmentation (PS) entails serializing bounding box coordinates and their object names as explained in \cref{fig:verbalization}. These verbalized descriptions are then transformed into auxiliary tokens through the word embeddings of MLM. Additionally, to directly utilize the panoptic segmentation map, we use a vision encoder and an MLP connector in MoAI to generate locality-preserving auxiliary tokens. The generated auxiliary tokens are flattened and concatenated to those from serialized bounding boxes and their object names to form the final PS auxiliary tokens $A_{\text{PS}}$. They are concatenated in this manner so that the MLM of MoAI can associate them in a compatible way through contextualization. This procedure ensures the comprehensive conversion of visual information from PS into language information while preserving the spatial locality inherent in the panoptic segmentation map. Note that if the panoptic segmentation model fails to classify objects within the fixed number of panoptic object categories, for instance, those in MS-COCO 2017~\cite{lin2014microsoft} encompassing 133 object categories, the \textit{unknown} class is assigned.

An open-world object detection model plays a role in detecting object classes missed by the panoptic segmentation model. This is because the panoptic segmentation model is trained on a specific dataset with a fixed number of object categories. Once the detection results are generated for an image, bounding box coordinates and their object names are verbalized according to the following template format: `The image includes bounding boxes and their objects: \{verbalized open-world object detection (OWOD) results\}'. Then, the results are transformed into OWOD auxiliary tokens $A_{\text{OWOD}}$ by the word embeddings of MLM. Similarly, the outputs of SGG and OCR models are verbalized, and corresponding auxiliary tokens $A_{\text{SGG}}$ and $A_{\text{OCR}}$ are generated, where we use the following verbalization templates: `The image includes relationships between objects: \{verbalized SGG results\}' and `The image includes text descriptions: \{verbalized OCR results\}', respectively.

\paragraph{\textbf{MoAI-Compressor.}} After the verbalization of CV model outputs, four auxiliary tokens $A_{\text{PS}}$, $A_{\text{OWOD}}$, $A_{\text{SGG}}$, and $A_{\text{OCR}}$ are generated and injected into MoAI-Compressor, which borrows the structure of Perceiver Resampler~\cite{alayrac2022flamingo}. All four auxiliary tokens $[A_{\text{PS}}, A_{\text{OWOD}}, A_{\text{SGG}}, A_{\text{OCR}}]$ are concatenated before being fed into MoAI-Compressor along with a fixed number of learnable tokens $A_{\text{input}}$, whose outputs $A$ are also fixed in length by the same number and represent the compressed and aligned auxiliary visual information, as formulated as follows:
\begin{equation}
    A = \text{MoAI-Compressor}(\left[A_{\text{PS}}, A_{\text{OWOD}}, A_{\text{SGG}}, A_{\text{OCR}}\right], A_{\text{input}}).
    \label{eqn:concat}
\end{equation}
Due to the variable length of concatenated auxiliary tokens across images and their substantial length after concatenation, MoAI-Compressor is designed to condense those tokens $[A_{\text{PS}}, A_{\text{OWOD}}, A_{\text{SGG}}, A_{\text{OCR}}]$ with a relatively small fixed size of 64, generating $A\in \mathbb{R}^{d\times64}$ where $d$ represents the embedding dimension. These condensed tokens are then used to extract relevant information for VL tasks by MoAI-Mixer. This compression enhances computational efficiency.

\begin{figure}[t!]
\vspace{-3mm}
    \centering
    \includegraphics[width=0.95\textwidth]{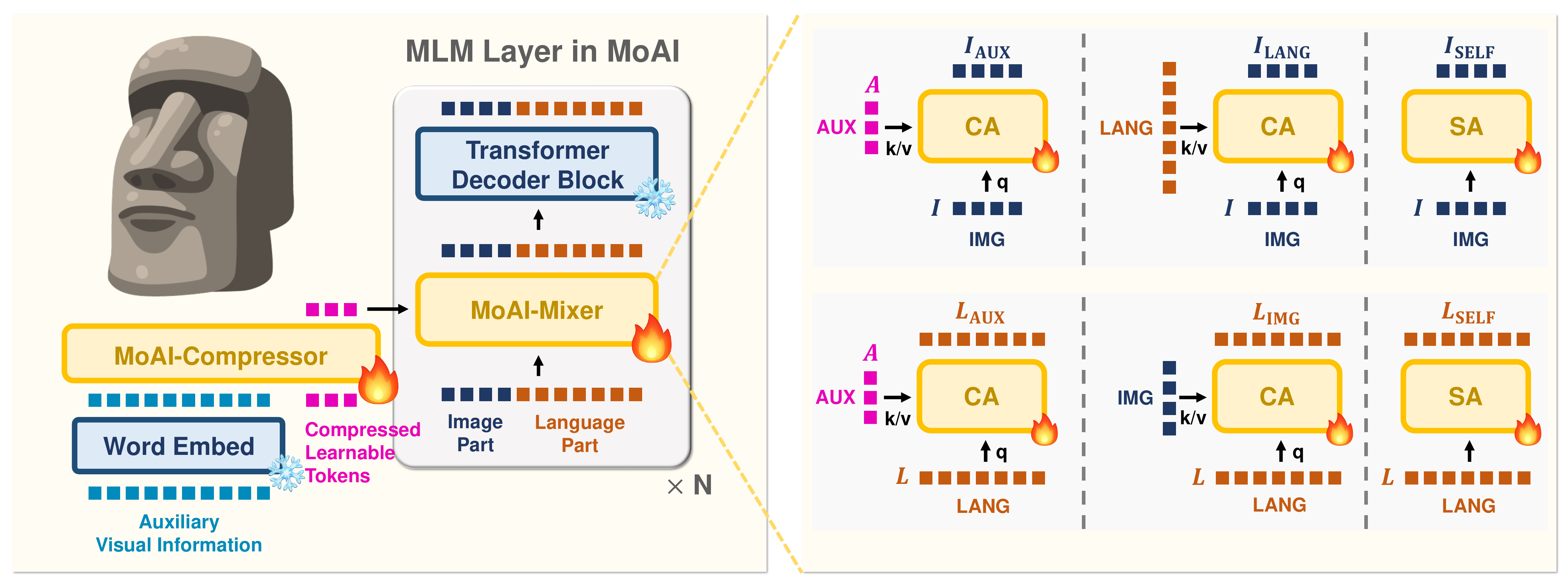}
    \vspace{-3mm}
    \caption{Illustrating MoAI-Mixer in MLM Layer of 
    \includegraphics[width=0.025\textwidth]{figure/moai.pdf} \textbf{MoAI}. In MoAI-Mixer, there are six expert modules to harmonize auxiliary features $A$ and two original features (\textit{i.e.,} visual $I$ and language $L$ features). }
    \label{fig:moai-mixer}
    \vspace{-2mm}
\end{figure}

\begin{figure}[t!]
\vspace{-3mm}
    \centering
    \includegraphics[width=0.9\textwidth]{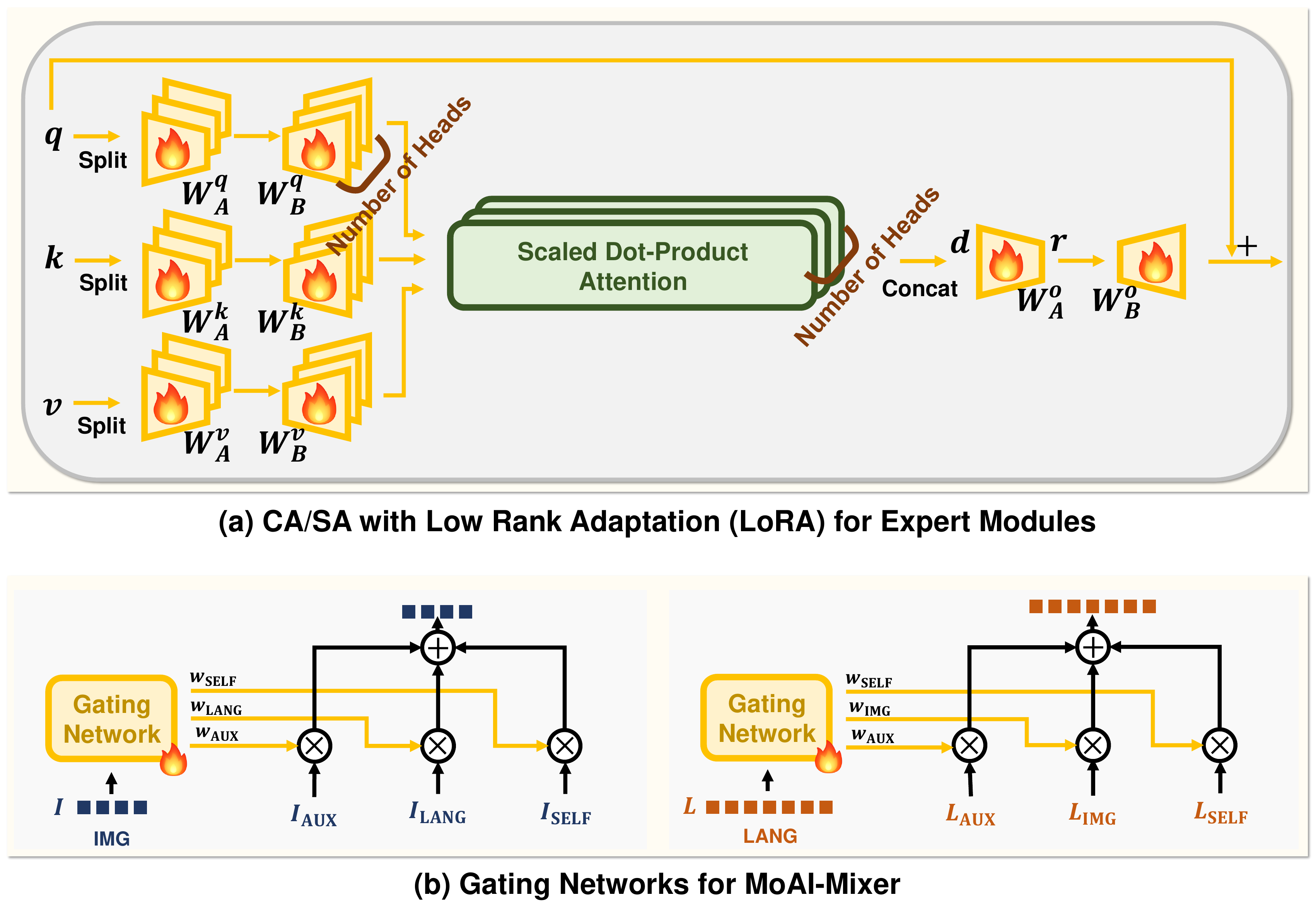}
    \vspace{-2mm}
    \caption{The structures of (a) expert modules and (b) gating networks for MoAI-Mixer. In (a), `$q$', `$k$', and `$v$' denote query, key, and value, respectively, `$d$' and `$r$' explains channel dimension and reduced dimension, respectively.}
    \label{fig:gating}
    \vspace{-5mm}
\end{figure}

\paragraph{\textbf{MoAI-Mixer}} is embedded in each MLM layer of MoAI. It receives auxiliary tokens $A$ from MoAI-Compressor, visual features $I^{(l)}\in\mathbb{R}^{d\times N_{I}}$, and language features $L^{(l)}\in\mathbb{R}^{d\times N_{L}}$ where $l=0,1,\cdots,N-1$ denotes the layer index, $d$ denotes the embedding dimension, $N_I$ denotes the length of visual features, and $N_L$ denotes that of language features. Normally, an MLM layer only consists of a Transformer decoder block $\text{TransDec}^{(l)}$ such that $[I^{(l+1)}, L^{(l+1)}]=\text{TransDec}^{(l)}([I^{(l)}, L^{(l)}])$. In MoAI, an $l$-th MLM layer with MoAI-Mixer is formulated as follows:
\begin{equation}
\begin{split}
    [\hat{I}^{(l)}, \hat{L}^{(l)}]&=\text{MoAI-Mixer}^{(l)}(A, I^{(l)}, L^{(l)}), \\\\
    [I^{(l+1)}, L^{(l+1)}]&=\text{TransDec}^{(l)}(\hat{I}^{(l)}, \hat{L}^{(l)}),
\end{split}
\label{eqn:moai-mixer}
\end{equation} 
where $\hat{I}^{(l)}$ and $\hat{L}^{(l)}$ are mixed visual features and mixed language features. In each MoAI-Mixer, we design six expert modules that are either cross- or self-attention modules as illustrated in \cref{fig:moai-mixer}: three for visual features $I$ and three for language features $L$. Each of three expert modules for visual features outputs $I_{\text{AUX}}$, $I_{\text{LANG}}$, and $I_{\text{SELF}}$ where the capital letter indicates query features and the subscript indicates key/value features. Similarly, each of three expert modules for language features outputs $L_{\text{AUX}}$, $L_{\text{IMG}}$, and $L_{\text{SELF}}$. The cross-attention operation at the $l$-th layer is formulated as follows:
\begin{equation}
\begin{split}
    I^{(l)}_{\text{\{AUX or LANG\}}}&=\text{CA}^{(l)}(q=I^{(l)}, k=\{A \text{ or } L^{(l)}\}, v=k), \\\\
    L^{(l)}_{\text{\{AUX or IMG\}}}&=\text{CA}^{(l)}(q=L^{(l)}, k=\{A \text{ or } I^{(l)}\}, v=k).
\end{split}
\label{eqn:moai-mixer-operation}
\end{equation}
In addition, the self-attention operation is formulated as $I^{(l)}_{\text{SELF}}=\text{SA}^{(l)}(I^{(l)})$ and $L^{(l)}_{\text{SELF}}=\text{SA}^{(l)}(L^{(l)})$. These six expert modules explicitly specialize in one of the following six distinct mixtures of intelligence: $I_{\text{AUX}}$, $I_{\text{LANG}}$, $I_{\text{SELF}}$, $L_{\text{AUX}}$, $L_{\text{IMG}}$, and $L_{\text{SELF}}$. When training the expert modules, we borrow the concept of LoRA~\cite{hu2021lora} to reduce computational burden. Let's denote $W$ as a general notation for a linear projection layer in a multi-head attention module~\cite{vaswani2017attention}, which can be $W^{q}$, $W^{k}$, $W^{v}$, or $W^{o}$. We decompose $W \in \mathbb{R}^{d\times d}$, not $\Delta W$ as in LoRA, into two linear layers $W_A \in\mathbb{R}^{d\times r}$ and $W_B \in\mathbb{R}^{r\times d}$ such that $W=W_{A} W_{B}$. The hyperparameter $r$ denotes the reduced dimension as illustrated in \cref{fig:gating}(a). Since computational burden of an attention module mainly comes from the high embedding dimension, usually $d=4096$, such formulation of projection matrices significantly reduces computation. Moreover, the input query features are directly added to the output features so that mixture of intelligence occurs without altering the outputs of the previous MLM layer too much, stabilizing the optimization process with the frozen Transformer decoder blocks.

\paragraph{\textbf{First Training Step.}} Together with the MLP connector, we first train $A_{\text{input}}$, MoAI-Compressor, and MoAI-Mixer by using visual instruction tuning datasets~\cite{liu2023improved, chen2023sharegpt4v}. This step ensures that the six expert modules in MoAI-Mixer yield meaningful features to conduct VL tasks. To do so, we randomly choose outputs from one of three expert modules for visual and language features, respectively, as follows:
\begin{equation}
    \hat{I}^{(l)}=\text{Sample}(I^{(l)}_{\text{AUX}},I^{(l)}_{\text{LANG}},I^{(l)}_{\text{SELF}}),\quad\hat{L}^{(l)}=\text{Sample}(L^{(l)}_{\text{AUX}},L^{(l)}_{\text{IMG}},L^{(l)}_{\text{SELF}}).
\label{eqn:moai-mixer-step1}
\end{equation}
Then, they are injected into the transformer decoder block $\text{TransDec}_l(\hat{I}^{(l)}, \hat{L}^{(l)})$. This sampling process aims for each expert module to produce meaningful features independently.

\paragraph{\textbf{Second Training Step.}} In this step, we extend the learning process beyond the parameters learned in the first training step. We learn two gating networks for each MoAI-Mixer, which comprises a single linear layer, each for visual and language features: $W_{\text{Gating}_I}$ and $ W_{\text{Gating}_L}\in\mathbb{R}^{d\times 3}$, illustrated in \cref{fig:gating}(b). 
The gating networks aim to output the best combination of weights for three expert modules for visual and language features each by using a linear layer and a softmax function as follows: $\text{Softmax}(x^{\mathsf{T}} W_{\text{Gating}_x}, \text{dim=1})$. Note that $x \in \mathbb{R}^{d \times N_x}$, where $x$ is either the visual $I$ or language $L$ features and $N_x$ is the length of features, resulting in $x^{\mathsf{T}} W_{\text{Gating}_x}\in\mathbb{R}^{N_x \times 3}$. Then, we split the softmax matrix into three weight vectors: $\text{Softmax}(x^{\mathsf{T}} W_{\text{Gating}_x}, \text{dim=1}) \rightarrow [w_{\text{AUX}}, w_{\text{LANG}}, w_{\text{SELF}}]$ where each weight has $\mathbb{R}^{N_x}$ dimension. The weights serve as confidence scores to determine whether to use information from each expert module. From the outputs of the gating networks, the propagation flow for the three sources of intelligence: `AUX', `IMG', `LANG' can be represented as follows:
\begin{equation}
    \begin{split}
        &[w_{\text{AUX}}, w_{\text{LANG}}, w_{\text{SELF}}] \leftarrow \text{Softmax}({I^{(l)}}^{\mathsf{T}} W_{\text{Gating}_I}, \text{dim=1}),\\\\
        &\hat{I}^{(l)}=w_{\text{AUX}}\odot I^{(l)}_{\text{AUX}}+w_{\text{LANG}}\odot I^{(l)}_{\text{LANG}}+w_{\text{SELF}}\odot I^{(l)}_{\text{SELF}}\\\\
        &[w_{\text{AUX}}, w_{\text{IMG}}, w_{\text{SELF}}] \leftarrow \text{Softmax}({L^{(l)}}^{\mathsf{T}} W_{\text{Gating}_L}, \text{dim=1}),\\\\
        &\hat{L}^{(l)}=w_{\text{AUX}}\odot L^{(l)}_{\text{AUX}}+w_{\text{IMG}}\odot L^{(l)}_{\text{IMG}}+w_{\text{SELF}}\odot L^{(l)}_{\text{SELF}},\\\\
    \end{split}
\end{equation}
where $\odot$ represents the element-wise product in each token. The gating networks for visual and language features are trained independently without parameter sharing, ensuring that both gating networks blend the three intelligence with different weights. In this manner, MoAI-Mixer and gating networks facilitate the interaction among the three sources of intelligence.

\section{Experiments}
\label{sec:experiment}

\paragraph{\textbf{Implementation Details.}} To ensure successful reproducibility, we outline three crucial technical details of MoAI: (a) external CV models, (b) MoAI-Compressor and MoAI-Mixer, (c) training and inference details.

\definecolor{Gray}{gray}{0.93}
\definecolor{Green}{rgb}{0.9, 0.95, 0.97}
\newcommand{\cmark}{\ding{51}}%
\newcommand{\xmark}{\ding{55}}%

\begin{table}[t!]
\vspace{-3mm}
\centering
\caption{Evaluating zero-shot performances of \includegraphics[width=0.025\textwidth]{figure/moai.pdf} \textbf{MoAI} on nine vision language datasets compared with the current powerful VLMs on Q-Bench~\cite{wu2023q}, SQA-IMG~\cite{iyyer2017search}, TextVQA~\cite{singh2019towards}, POPE~\cite{li2023evaluating}, MME(-P, -C)~\cite{fu2023mme}, MM-Bench(-CN)~\cite{liu2023mmbench}, and MM-Vet~\cite{yu2023mm}.}
\vspace{-2mm}
\label{tab:zeroshot}
\resizebox{\linewidth}{!}{
\renewcommand{\tabcolsep}{1mm}
\begin{tabular}{lcccccccccccc}
\toprule
VLMs             & Q-Bench & SQA-IMG   & TextVQA & POPE & MME-P & MME-C & MM-Bench & MMB-CN         & MM-Vet \\
\midrule
BLIP2-13B~\cite{blip2}       & -       & 61.0      & 42.5    & 85.3 & 1294& 290 & -        & -              & 22.4   \\
\rowcolor{Gray}
InstructBLIP-7B~\cite{dai2023instructblip} & 56.7    & 60.5      & 50.1    & -    & -     & -     & 36.0       & 23.7           & 26.2   \\
\rowcolor{Gray}
InstructBLIP-13B~\cite{dai2023instructblip} & -       & 63.1      & 50.7    & 78.9 & 1213& -     & -        & -              & 25.6   \\
Shikra-13B~\cite{chen2023shikra}      & 54.7    & -         & -       & -    & -     & -     & 58.8     & -              & -      \\
IDEFICS-9B~\cite{laurenccon2023obelisc}      & -       & -         & 25.9    & -    & -     & -     & 48.2     & 25.2           & -      \\
IDEFICS-80B~\cite{laurenccon2023obelisc}     & -       & -         & 30.9    & -    & -     & -     & 54.5     & 38.1           & -      \\
\rowcolor{Gray}
Qwen-VL-7B~\cite{bai2023qwen}      & 59.4    & 67.1      & 63.8    & -    & -     & -     & 38.2     & 7.4            & -      \\
\rowcolor{Gray}
Qwen-VL-Chat-7B~\cite{bai2023qwen} & -       & 68.2      & 61.5    & -    & 1488& 361 & 60.6     & 56.7           & -      \\
MiniGPT-4-7B~\cite{zhu2023minigpt}    & -       & -         & -       & -    & 582 & -     & 23.0     & -              & 22.1   \\
Otter-7B~\cite{li2023otter}        & 47.2    & -         & -       & -    & 1292& -     & 48.3     & -              & 24.6   \\
LLaVA-7B~\cite{liu2023visual}        & -       & 38.5      & -       & -    & 807 & 248 & 34.1     & 14.1           & 26.7   \\
MiniGPT-v2-7B~\cite{chen2023minigpt}   & -       & -         & -       & -    & -     & -     & -        & -              & -      \\
MiniGPT-v2-Chat-7B~\cite{chen2023minigpt} & -       & -         & -       & -    & -     & -     & -        & -              & -      \\
\rowcolor{Gray}
LLaVA1.5-7B~\cite{liu2023improved}     & 58.7    & 66.8      & 58.2    & {85.9} & 1511& 294& 64.3     & 58.3           & 30.5   \\
\rowcolor{Gray}
LLaVA1.5-13B~\cite{liu2023improved}    & 62.1    & {71.6}      & {61.3}    & {85.9} & 1531& 295 & 67.7     & 63.6           & 35.4   \\
mPLUG-Owl-7B~\cite{ye2023mplug}    & 58.9    & -         & -       & -    & 967 & -     & 46.6     & -              & -      \\
mPLUG-Owl2-7B~\cite{ye2023mplug2}   & 62.9    & 68.7      & 58.2    &      & 1450& -     & 64.5     & -              & 36.2   \\
ShareGPT4V-7B~\cite{chen2023sharegpt4v}   & 63.4    & 68.4      & -       &      & {1567}& 376 & 68.8     & 62.2              & {37.6}   \\
CogVLM-17B~\cite{wang2023cogvlm}      & -       & 68.7      & 58.2    &      & -     & -     & 65.8     & 55.9           & \textbf{54.5}   \\
LLaVA-XTuner-20B~\cite{2023xtuner}    & -       & -         & -       & -    & -     & -     & {75.1}     & {73.7}           & 37.2   \\
Intern-XC-7B~\cite{zhang2023internlm} & 64.4   & -         & -       &      & 1528& {391} & 74.4     & 72.4           & 35.2   \\
\midrule
\rowcolor{Green} 
MoAI-7B      & \textbf{70.2}   & \textbf{83.5}      & \textbf{67.8}    &  \textbf{87.1} & \textbf{1714}& \textbf{561}& \textbf{79.3}    & \textbf{76.5}           & 43.7   \\
\bottomrule 
\end{tabular}
}
\vspace{-4mm}
\end{table}

\paragraph{\textbf{(a)}} For panoptic segmentation, we adopt Mask2Former~\cite{cheng2022masked} (model size: 106M) with Swin-B/4~\cite{liu2021swin}. To predict a panoptic segmentation map, we set the threshold to keep predicted instance masks as $0.5$ and set the mask threshold to use the masks as $0.95$. For open-world object detection, we use OWLv2~\cite{minderer2023scaling} (model size: 154M) with CLIP-B/16~\cite{clip}. To achieve open-world object detection, we deal with 1847 object categories combining those in ADE20K-847~\cite{zhou2017scene, zhou2019semantic} and ImageNet~\cite{deng2009imagenet}. We set the threshold to keep object detection predictions as $0.1$ and set the object threshold to use them as $0.5$. For scene graph generation (SGG), we utilize panoptic SGG~\cite{yang2022panoptic} (model size: 44M) with ResNet-50~\cite{he2016deep} to conduct flexible interactions with foreground and background objects, where $0.8$ threshold to use SGG predicates is set. For OCR, we use PaddleOCRv2~\cite{du2021pp} (model size: 18M), one of performant open-source OCR frameworks, where we set recognizable languages to Chinese \& English and set hyper-parameter settings to possibly read rotated text descriptions. The combined size of the external CV models is about 332M, contributing a little to the total model size.

\paragraph{\textbf{(b)}} In MoAI-Compressor, the learnable tokens $A_{\text{input}}$ have $\mathbb{R}^{4096\times64}$ dimension where $64$ denotes the number of tokens (length) and $4096$ represents the channel dimension $d$ for MLM input. In addition, MoAI-Compressor comprises $4$ standard Transformer encoder layers~\cite{vaswani2017attention}. In the self-attention, $4$ number of heads and $64$ head dimension are set. To build MoAI-Mixer, we equip it with specific MLM layer indices $l=7, 15, 23, 31$. For CA/SA expert modules, $64$ reduced dimension, $4$ number of heads, and $4096/4=1024$ head dimension are used.

\paragraph{\textbf{(c)}} For all training steps, we deal with a standard visual instruction tuning dataset: LLaVA-Instruct-665K~\cite{liu2023improved} filtered by \cite{chen2023sharegpt4v}. Regarding the first training step, we train the learnable tokens $A_{\text{input}}$, the parameters of MoAI-Compressor, and six expert modules of MoAI-Mixer in one epoch using the AdamW~\cite{loshchilov2018decoupled} optimizer, scheduled by cosine annealing~\cite{loshchilov2016sgdr} from learning rate of 1e-4 to 1e-6. In the second training step, we not only learn the parameters trained in the first training step but also the gating networks, where learning rate is scheduled from 2e-5 to 1e-6 in one epoch. For efficient inference, we quantize MoAI in 4-bit where double quantization and normalized float 4-bit (nf4)~\cite{dettmers2023qlora} are used, and we use deterministic beam search ($n=3$)~\cite{freitag-al-onaizan-2017-beam} for text generation.


\begin{table}[t!]
\vspace{-3mm}
\centering
\caption{Illustrating the effectiveness of external computer vision (CV) models compared by the perception scores in MME~\cite{fu2023mme} and MM-Bench~\cite{liu2023mmbench}. `TT' denotes text translation task that requires OCR as a priority.}
\label{tab:ablation1}
\resizebox{\linewidth}{!}{
\renewcommand{\tabcolsep}{1mm}
\begin{tabular}{cccccccccccc}
\toprule
        &     &     & \multicolumn{5}{c}{MME} & \multicolumn{4}{c}{MM-Bench}                   \\
\cmidrule(lr){4-8}\cmidrule(lr){9-12}
PS+OWOD & SGG & OCR & Existence & Position & Scene & OCR & TT  & Recognition & Localization & Spatial & OCR\\
\midrule
\xmark       & \cmark   & \cmark   & 187   & 154& 161   & 145 & 138 & 77.6 & 54.0 & 32.6 & 84.6\\
\cdashline{1-12}\noalign{\vskip 0.5ex}
\cmark       & \xmark   & \cmark   & 198   & 145& 164   & 147 & 150 & 89.7 & 65.3 & 35.8 & 90.9 \\
\cdashline{1-12}\noalign{\vskip 0.5ex}
\cmark       & \cmark   & \xmark   & 199   & 163& 166   & 120 & 95  & 91.8 & 69.2 & 42.8 & 80.1\\
\cdashline{1-12}\noalign{\vskip 0.5ex}
\rowcolor{Green} 
\cmark       & \cmark   & \cmark   & \textbf{200}   & \textbf{165}      & \textbf{170}   & \textbf{148} & \textbf{153} & \textbf{92.9} & \textbf{71.1} & \textbf{43.2} & \textbf{93.5}\\
\bottomrule
\end{tabular}
}
\end{table}
\begin{table}[t!]
\caption{Ablation study for training step choice, selecting top-$k$ expert modules in MoAI-Mixer, and the type of weights for gating network.}
\vspace{-3mm}
\label{table:ablation2}
\centering
\begin{minipage}[t]{0.32\linewidth}
\caption*{(a) Training step choice}
\vspace{-2mm}
\centering
\resizebox{\linewidth}{!}{
\renewcommand{\tabcolsep}{1.5mm}
\begin{tabular}{lccccc}
\toprule
Step     & MME-P  & MME-C \\
\midrule
First    & 1542 & 369  \\
Second   & 1654 & 511  \\
\rowcolor{Green} 
Combined & \textbf{1714} & \textbf{561}  \\
\bottomrule
\end{tabular}
}

\end{minipage}
\begin{minipage}[t]{0.32\linewidth}
\caption*{(b) Selecting Top-$k$ Experts} 
\vspace{-2mm}
\centering
\resizebox{\linewidth}{!}{
\renewcommand{\tabcolsep}{3.5mm}
\begin{tabular}{lccccc}
\toprule
$k$ & MME-P  & MME-C\\
\midrule
1 & 1588 & 387  \\
2 & 1638 & 451 \\
\rowcolor{Green} 
3 & \textbf{1714} & \textbf{561}  \\
\bottomrule
\end{tabular}
}
\end{minipage}
\begin{minipage}[t]{0.32\linewidth}
\caption*{(c) Gating network weights}
\vspace{-2mm}
\centering
\resizebox{\linewidth}{!}{
\renewcommand{\tabcolsep}{1.8mm}
\begin{tabular}{lccccc}
\toprule
Gating  & MME-P  & MME-C \\
\midrule
Random  & 1520 & 348  \\
Uniform & 1617 & 485  \\
\rowcolor{Green} 
Trained & \textbf{1714} & \textbf{561}  \\
\bottomrule
\end{tabular}
}

\end{minipage}
\vspace{-4mm}
\end{table}

\paragraph{\textbf{Evaluating Visual Perception Capability.}} 
Delving into validating the effectiveness of MoAI, we look deeper into visual perception capability related to real-world scene understanding in numerous VL benchmarks, such as MME, SEED, MM-Bench, and MM-Vet. \cref{fig:perception} illustrates the zero-shot performances in detail of MoAI and three state-of-the-art open-source LLVMs such as InstructBLIP~\cite{dai2023instructblip}, Qwen-VL~\cite{bai2023qwen}, LLaVA1.5~\cite{liu2023improved}. For each VL benchmark, there exist specific dimensions (sub-benchmarks) related to real-world scene understanding in which MoAI aims to demonstrate its efficacy. Refer to \cref{app:A} for more details on what each dimension specifically indicates. As it can be seen from \cref{fig:perception}, MoAI significantly surpasses other LLVMs, demonstrating the effectiveness of utilizing auxiliary visual information from external CV models. It is noteworthy that MoAI especially excels at relation and text-related dimensions, emphasizing the significance of using auxiliary visual information that they struggle to fully comprehend. Refer to \cref{app:D} for qualitative assessment with demonstration on a few samples. Furthermore, \cref{tab:zeroshot} exhibits thorough evaluation across numerous renowned VL benchmarks, and demonstrates the exceptional performance of MoAI. The versatility of MoAI corroborates that enhancing real-world scene understanding can boost not only visual perception related to it but also overall VL capabilities in \cref{fig:llvms_open_close}(b).

\paragraph{\textbf{Ablation Studies.}} 
To validate the effectiveness of the external CV models we utilize, we conduct evaluation by subtracting them one by one. ~\cref{tab:ablation1} shows significant drop of object existence and recognition without using panoptic segmentation (PS) and open-world object detection (OWOD). On the other hand, once SGG is not used, the scores related with relations such as Position and Spatial are dropped in ~\cref{tab:ablation1}. In addition, the OCR scores are also dropped if OCR is not employed. Therefore, we can say that each of the external CV models is crucial for real-world scene understanding based on the perception scores for MME, SEED, MM-Bench, and MM-Vet. Additionally, we control three factors of MoAI-Mixer and gating networks in \cref{table:ablation2}: (a) the two training steps, (b) selecting top-$k$ in expert modules, and (c) weights of gating networks, in order to validate their effectiveness. Note that, selecting top-$k$ is based on the weight magnitude among $w_{\text{AUX}}$, $w_{\text{IMG}}$, $w_{\text{LANG}}$, and $w_{\text{SELF}}$, as higher weights signify greater importance of the associated information. Further, \cref{app:B} shows additional experiments for ablation studies.

\begin{figure}[t!]
\vspace{-3mm}
    \centering
    \includegraphics[width=0.95\textwidth]{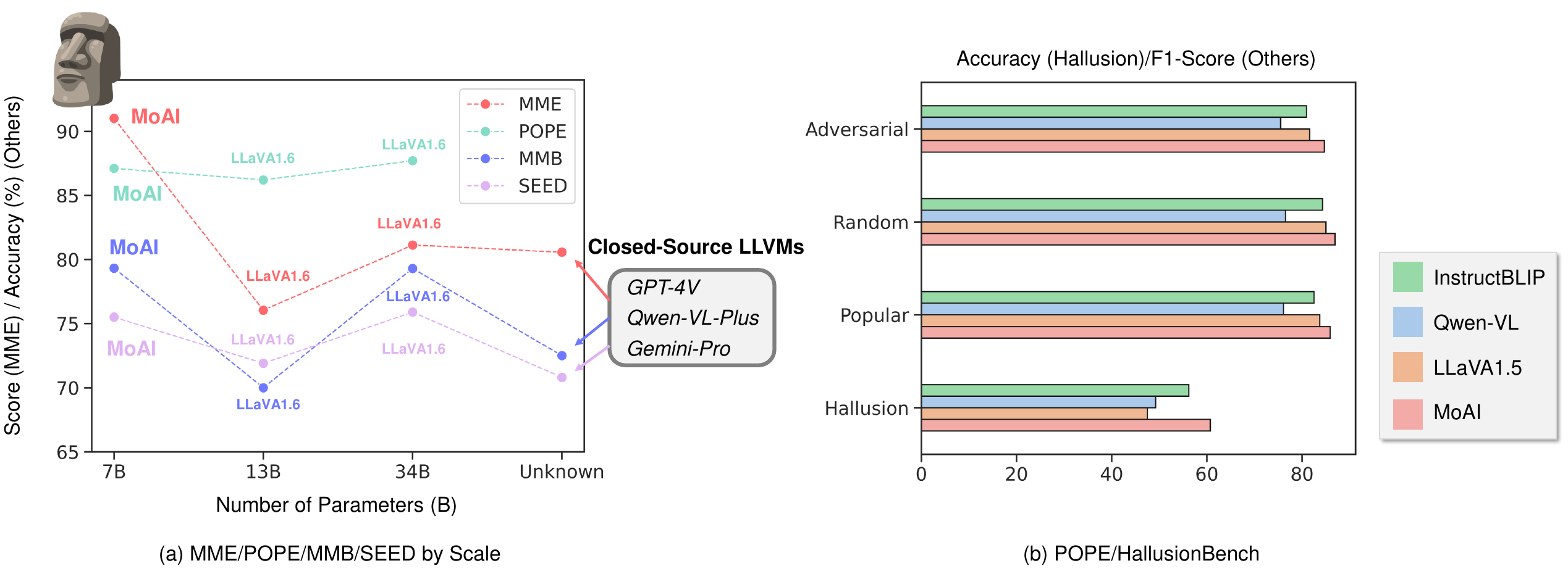}
    \vspace{-3mm}
    \caption{Illustrating zero-shot vision language performances (a) by model size scale compared with the larger open-source LLVMs: LLaVA1.6-13B and -34B~\cite{liu2024llavanext}, in the latest, and closed-source LLVMs. (b) shows the results of POPE~\cite{li2023evaluating} and HallusionBench~\cite{liu2023hallusionbench}, where `Adversarial', `Random', and `Popular' are metrics in POPE. Note that, the scores of MME in (a) are scaled down by 1/25 times to fit the figure, and the dot points for closed-source LLVMs represent averaged performances with them.}
    \label{fig:final}
    \vspace{-3mm}
\end{figure}

\paragraph{\textbf{Discussion.}}
From the results, we can obtain an insight that prioritizing real-world scene understanding is more crucial than relying on the extra curation of visual instruction datasets or scaling up model size. As illustrated in \cref{fig:final}(a), MoAI-7B surpasses the zero-shot performances, despite being relatively small compared to the considerably larger open-source and closed-source models. Notably, \cref{fig:final}(b) also indicates that MoAI performs well even  on hallucination zero-shot datasets: POPE~\cite{li2023evaluating} and HallusionBench~\cite{liu2023hallusionbench}. This suggests that accurately recognizing objects and their relationships can help prevent LLVMs from making mistakes. Looking ahead, as MoAI is tailored for real-world scene understanding, we plan to incorporate more external CV models to provide LLVMs with diverse capabilities for low-level vision understanding, common-sense knowledge, and awareness of non-object notions beyond text descriptions, such as charts, diagrams, signs, and symbols, as well as solving advanced math problems. Furthermore, robust~\cite{lee2020training, lee2020towards, lee2022masking, kim2023demystifying}, unbiased~\cite{liu2022show, lee2023mitigating, kim2023mitigating}, and explainable~\cite{kim2021distilling, caron2021emerging, kim2023causal} CV models can be applied to achieve precise and unbiased outputs for vision language tasks. Further, \cref{app:C} describes further discussion.

\section{Conclusion}
\label{sec:conclusion}
To achieve real-world scene understanding, we leverage  fundamental perception capabilities rooted in cognitive science and machine learning. This involves incorporating auxiliary visual information from historically rich external CV models, which we seemlessly integrate with visual and language features in MLM using expert modules and gating networks. As a result of these advancements, \includegraphics[width=0.025\textwidth]{figure/moai.pdf} MoAI demonstrates  improved visual perception capabilities, resulting in significant enhancements in zero-shot vision language performances. This underscores MoAI's potential to advance LLVM modeling by effectively leveraging diverse auxiliary visual information and integrating multiple forms of intelligence.

\clearpage  
\section*{Acknowledgements}
This work was partially supported by two funds: IITP grant funded by the Korea government (MSIT) (RS-2022-II220984) and Center for Applied Research in Artificial Intelligence (CARAI) grant funded by DAPA and ADD (UD230017TD). Additionally, it was supported by the KISTI National Supercomputing Center with supercomputing resources including technical support (KSC-2024-CRE-0160). 

%
%
\bibliographystyle{ref}
\bibliography{ref}

\clearpage
\appendix

\section{Details for Specific Dimensions in \cref{fig:perception}}
\label{app:A}
This section explains details on dimensions (sub-benchmarks or sub-tasks) related to real-world scene understanding in the following benchmarks: MME, SEED, MM-Bench, and MM-Vet.


\subsection{MME}
\begin{itemize}
    \item \textbf{Existence} pertains to inquiries concerning the presence of a single object within a specified image.
    \item \textbf{Count} denotes the process of quantifying instances of the specified object depicted in an image.
    \item \textbf{Position} describes the capacity of a model to discern the spatial arrangement between two objects within a given image.
    \item \textbf{Scene} focuses on the identification of a place shown in an image, including indoor and outdoor locations such as a gallery, laboratory, lakeside, landmark, etc.
    \item \textbf{OCR} serves to measure the capability of a model to recognize texts in the image.
    \item \textbf{Text translation} requires a model, supporting both English and Chinese, to translate the Chinese script in an image to the corresponding English.
    
\end{itemize}

\subsection{SEED}
\begin{itemize}
    \item \textbf{Scene (scene understanding)} focuses on the overall information conveyed in the image, where inquiries can be addressed through a comprehensive grasp of the image's content.
    \item \textbf{Identity (instance identity)} involves the object recognition capability of a model, which includes determining the presence or categorization of an instance.
    \item \textbf{Attributes (instance attributes)} relates to the distinguishing features of an instance, such as its color, shape, or material composition, serving to evaluate the model's comprehension of an object's visual appearance.
    \item \textbf{Location (instance location)} concerns the precise spatial coordinates of a specific instance within the image, necessitating accurate localization of the referenced object by the model.
    \item \textbf{Count (instances counting)} pertains to the capability of a model to count the occurrences of a designated object within the image, demanding a comprehensive understanding and enumeration of instances of the specified object.
    \item \textbf{Relation (spatial relation)} prompts the model to establish the spatial connection between two specified objects within the image, grounding them and recognizing their relative spatial arrangement.
    \item \textbf{Interaction (instance interaction)} requires the model to identify either the relational state or interactive dynamics between two human entities or objects depicted in the image.
    \item \textbf{Text recognition (text understanding)} involves inquiries concerning the textual components present within the image, prompting the model to comprehend and interpret textual elements accordingly.
    
\end{itemize}

\subsection{MM-Bench}
\begin{itemize}
    \item \textbf{Attribute (attribute recognition)} involve the capability of a model to recognize various features of an image such as texture, shape, appearance, emotions, categories, celebrities, renowned locations, objects, and optical characters within an image.
    \item \textbf{Localization (object localization)} involves inquiries regarding an object's position, absolute coordinates, count, and orientation within the image.
    \item \textbf{OCR} showcases a model's capability to recognize text, formulas, and sheets within an image.
    \item \textbf{Relation (spatial relationship)} assesses the capability of a model to understand the relative positions between objects depicted in the image.
    
\end{itemize}

\subsection{MM-Vet}
\begin{itemize}
    \item \textbf{Recognition}  evaluates the overall visual recognition proficiency of a model, encompassing the identification of scenes, objects, object attributes, counting, and various other advanced visual recognition tasks in computer vision.
    \item \textbf{Spatial (spatial awareness)} encompasses a wide range of abilities related to spatial comprehension, including understanding the spatial relationships among objects and textual regions within a scene.
    \item \textbf{OCR} assesses the capability of a model to understand and reason over scene text, where the model is evaluated on its capability to read text within images and utilize this information to solve various tasks.
    
\end{itemize}

\section{Further Experiments}
\label{app:B}

\paragraph{\textbf{Auxiliary Information on Other Baselines.}} In the following table, we have experimented simple ablation study to identify the effectiveness of auxiliary information, and then we observed the same increased results.
\begin{table}[h!]
\vspace{-7mm}
\resizebox{\linewidth}{!}{
\renewcommand{\tabcolsep}{4mm}
\begin{tabular}{lcccc}
\toprule
Dataset  & Qwen-VL-Chat-7B & \textbf{+Aux} & LLaVA1.5-7B & \textbf{+Aux} \\
\midrule
TextVQA  & 63.8         & \textbf{69.2} & 58.2     & \textbf{65.6} \\
\cdashline{1-5}\noalign{\vskip 0.5ex}
MME      & 1849         & \textbf{2192} & 1805     & \textbf{2187} \\
\cdashline{1-5}\noalign{\vskip 0.5ex}
MM-Bench & 60.6         & \textbf{77.5} & 64.3     & \textbf{79.1} \\
\bottomrule
\end{tabular}
}
\vspace{-5mm}
\end{table}

\paragraph{\textbf{Over-Complexity.}} We additionally validated more thorough ablation study in the following table. From these results, we can say that adding bounding box or segmentation map directly affects the sub-benchmarks for position, localization, spatial relation compared existence and recognition. They are hugely related with fine-grained understanding to objects. From these observations, we would like to claim that we meticulously selected external computer vision models and their output verbalization prompts, so all we used is necessary to enhance perception capabilities in balance. Nonetheless, open-world segmentation might be a solution to integrating segmentation and open-world detection, but, to the best of our knowledge, it has complicated structures having larger model sizes. This is why we considered the two independently in the current version.
\begin{table}[h!]
\vspace{-4mm}
\resizebox{\linewidth}{!}{
\renewcommand{\tabcolsep}{5mm}
\begin{tabular}{ccccccc}
\toprule
Box & Seg & Exist & Pos & Recog & Loc & Spatial \\
\midrule
\xmark   & \xmark   & 198       & 155      & 90.1        & 55.2         & 34.5                 \\
\cdashline{1-7}\noalign{\vskip 0.5ex}
\cmark   & \xmark   & 199       & 162      & 92.2        & 68.3         & 39.0                 \\
\cdashline{1-7}\noalign{\vskip 0.5ex}
\xmark   & \cmark   & 199       & 159      & 91.5        & 56.8         & 35.4                 \\
\cdashline{1-7}\noalign{\vskip 0.5ex}
\cmark   & \cmark   & \textbf{200}       & \textbf{165}      & \textbf{92.9}        & \textbf{71.1}         & \textbf{43.2}                 \\
\bottomrule
\end{tabular}
}
\vspace{-5mm}
\end{table}

\paragraph{\textbf{Inference Latency.}} We measured the number of generated tokens per second in Qwen-VL (16 tok/s), LLaVA1.5 (22 tok/s), and MoAI (20 tok/s) under flash attention in equal resource environments: Intel(R) Xeon(R) Gold 6230, RAM 512GB, and NVIDIA RTX A6000. Despite using external models, their size is relatively small and generation speed is highly related with multimodal language model size, so they did not affect inference speed much. This is similar to SAM~\cite{Kirillov_2023_ICCV} decoding process (encoded features are saved).

\paragraph{\textbf{Number of Classes.}} Below, we conducted simple ablation study to show considering total 1847 categories is enough by randomly choosing categories with 20 repetitive iterations.
\begin{table}[h!]
\vspace{-4mm}
\resizebox{\linewidth}{!}{
\renewcommand{\tabcolsep}{3mm}
\begin{tabular}{lccccccc}
\toprule
Datasets & 0    & 100  & 200  & 400  & 800  & 1200 & 1847 \\
\midrule
MME      & 1814 & 1955 & 2064 & 2172 & 2270 & \textbf{2275} & \textbf{2275} \\
\cdashline{1-8}\noalign{\vskip 0.5ex}
MM-Bench & 65.9 & 73.5 & 76.8 & 79.2 & 79.2 & \textbf{79.3} & \textbf{79.3} \\
\cdashline{1-8}\noalign{\vskip 0.5ex}
MM-Vet   & 33.2 & 38.1 & 42.0 & 43.6 & \textbf{43.7} & \textbf{43.7} & \textbf{43.7} \\
\bottomrule
\end{tabular}
}
\vspace{-5mm}
\end{table}

\newpage
\section{Further Discussion}
\label{app:C}

\paragraph{\textbf{Potential Dataset Biases.}} We evaluated \includegraphics[width=0.025\textwidth]{figure/moai.pdf} MoAI on MMStar~\cite{chen2024we} to handle potential dataset biases: some questions may not require visual information for accurate responses. From this result, we can say that MoAI overcomes the biases well compared with other baselines.
\begin{table}[h!]
\vspace{-3mm}
\resizebox{\linewidth}{!}{
\renewcommand{\tabcolsep}{3mm}
\begin{tabular}{lccccccc}
\toprule
Models          & CP   & FP   & IR   & LR   & ST   & MA   & Avg  \\
\midrule
LLaVA1.5-7B     & 58.8 & 24.0 & 38.8 & 24.0 & 13.6 & 22.8 & 30.3 \\
\cdashline{1-8}\noalign{\vskip 0.5ex}
Qwen-VL-Chat-7B & 59.6 & 32.0 & 50.8 & 29.2 & 22.0 & 31.6 & 37.5 \\
\cdashline{1-8}\noalign{\vskip 0.5ex}
MoAI-7B         & \textbf{68.5} & \textbf{46.3} & \textbf{59.1} & \textbf{42.9} & \textbf{35.7} & \textbf{40.2} & \textbf{48.7} \\
\bottomrule
\end{tabular}
}
\vspace{-4mm}
\end{table}

\paragraph{\textbf{Limited Performances for Computer Vision Models}} As we all know, there are no perfect models existing in the world that provide 100\% prediction accuracy. Consequently, inherent limitations such as erroneous predictions and biases from training datasets are inevitable. Nonetheless, despite their shortcomings, external models have demonstrated a high potential to enhance the perception capabilities of LLVMs, which previously struggled with these issues.

\paragraph{\textbf{Future Works.}} Despite its advanced performances for fundamental perception capabilities, \includegraphics[width=0.025\textwidth]{figure/moai.pdf} MoAI still requires low-level vision understanding, common-sense knowledge, charts, diagrams, signs, and symbols, advanced math problems, which are all beyond perception~\cite{lee2024meteor} with computer vision models. Therefore, we should extend the literal intelligence derived from additional computer vision models to encompass multifaceted information that includes fundamental image understanding, real-world common-sense knowledge, and non-object concepts. This expansion should cover charts, diagrams, symbols, signs, math problems, and step-by-step procedures for solving complex questions. Besides, we are struggling with building advanced architectures~\cite{lee2024trol} to embed more rich vision language knowledge, stimulating the effect of blending numerous intelligence.

\newpage
\section{Demonstrating \includegraphics[width=0.035\textwidth]{figure/moai.pdf} MoAI for Qualitative Assessment}
\label{app:D}
\begin{table}[h!]
\vspace{-6mm}
\centering
\resizebox{\linewidth}{!}{
\renewcommand{\tabcolsep}{1.5mm}
\begin{tabular}{ll}
\toprule
\multicolumn{2}{l}{SEED Visual input example, Scene Understanding:} \\
\midrule
    & \includegraphics[scale=0.45]{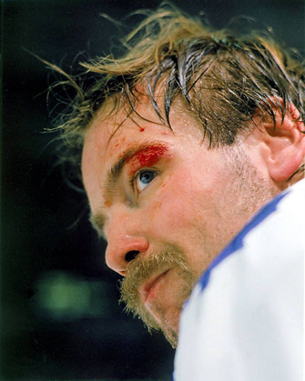} \\
User    & Which player in the image is bleeding from the face? \\
 & A. The hockey player B. The man with the beard \\
 & C. The tennis player D. None of them are bleeding\\
\midrule
InstructBLIP   & A. The hockey player \\
\midrule
Qwen-VL & A. The hockey player \\
\midrule
LLaVA1.5 & A. The hockey player \\
\midrule
MoAI & B. The man with the beard \\
\bottomrule
\end{tabular}
}
\end{table}
\begin{table}[h!]
\vspace{-5mm}
\centering
\resizebox{\linewidth}{!}{
\small
\renewcommand{\tabcolsep}{1.5mm}
\begin{tabular}{ll}
\toprule
\multicolumn{2}{l}{SEED Visual input example, Instance Identity:} \\
\midrule
    & \includegraphics[scale=0.3]{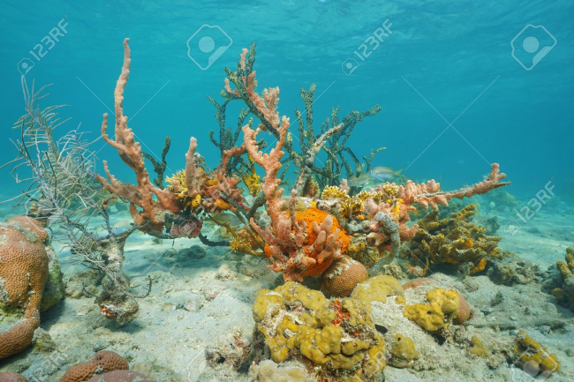} \\
User    & Which of the following is not present in the image? \\
 & A. Coral reef B. Palm tree C. Seashell D. Fish\\
\midrule
InstructBLIP   & D. Fish \\
\midrule
Qwen-VL & D. Fish \\
\midrule
LLaVA1.5 & D. Fish \\
\midrule
MoAI & B. Palm Tree \\
\bottomrule
\end{tabular}
}
\end{table}
\newpage
\begin{table}[h!]
\centering
\resizebox{\linewidth}{!}{
\renewcommand{\tabcolsep}{1.5mm}
\begin{tabular}{ll}
\toprule
\multicolumn{2}{l}{SEED Visual input example, Instance Attribute:} \\
\midrule
    & \includegraphics[scale=0.35]{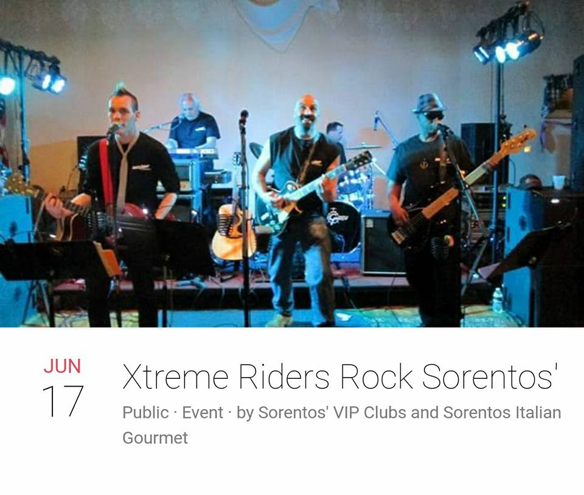} \\
User    & What is the action of the person wearing the hat on stage? \\
 & A. Playing the guitar B. Singing C. Dancing D. Playing the drums\\
\midrule
InstructBLIP   & A. Playing the guitar \\
\midrule
Qwen-VL & D. Playing the drums \\
\midrule
LLaVA1.5 & D. Playing the drums \\
\midrule
MoAI & A. Playing the guitar \\
\bottomrule
\end{tabular}
}
\end{table}
\begin{table}[h!]
\vspace{-9mm}
\centering
\resizebox{\linewidth}{!}{
\renewcommand{\tabcolsep}{5mm}
\begin{tabular}{l p{7cm}}
\toprule
\multicolumn{2}{l}{SEED Visual input example, Instance Location:} \\
\midrule
    & \includegraphics[scale=0.22]{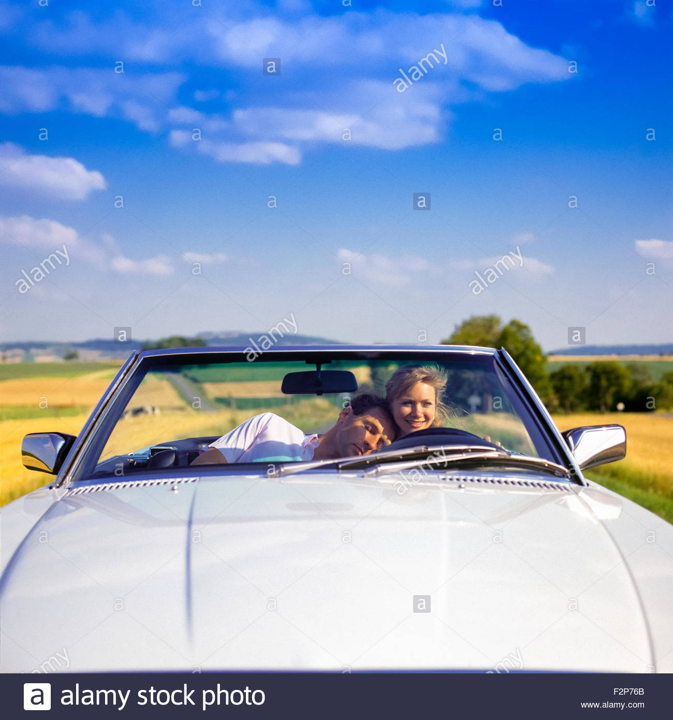} \\
User    & Where is the couple in the image located? \\
 & A. Behind the car B. In front of the car C. Inside the car D. On top of the car\\
\midrule
InstructBLIP   & C. Inside the car \\
\midrule
Qwen-VL & B. In front of the car \\
\midrule
LLaVA1.5 & D. On top of the car  \\
\midrule
MoAI & C. Inside the car \\
\bottomrule
\end{tabular}
}
\end{table}
\newpage
\begin{table}[h!]
\centering
\resizebox{\linewidth}{!}{
\renewcommand{\tabcolsep}{2.5mm}
\begin{tabular}{ll}
\toprule
\multicolumn{2}{l}{SEED Visual input example, Instance Counting:} \\
\midrule
    & \includegraphics[scale=0.28]{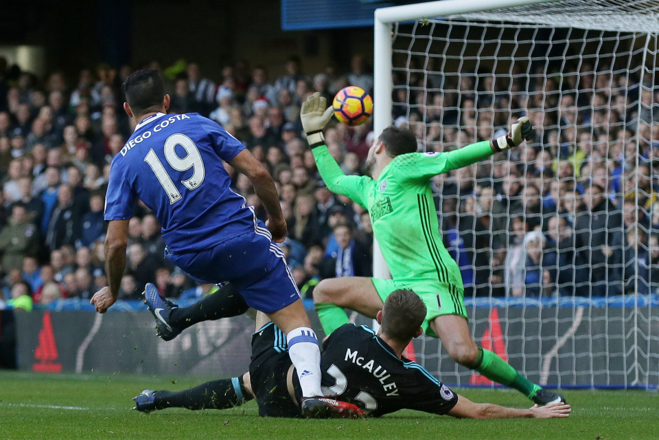} \\
User    & How many soccer players are present in the image? \\
 & A. 3  B. 2  C. 1  D. 4\\
\midrule
InstructBLIP   & D. 4 \\
\midrule
Qwen-VL & D. 4 \\
\midrule
LLaVA1.5 & D. 4 \\
\midrule
MoAI & A. 3 \\
\bottomrule
\end{tabular}
}
\end{table}
\begin{table}[h!]
\vspace{-7mm}
\centering
\resizebox{\linewidth}{!}{
\renewcommand{\tabcolsep}{4mm}
\begin{tabular}{l p{7cm}}
\toprule
\multicolumn{2}{l}{SEED Visual input example, Spatial Relation:} \\
\midrule
    & \includegraphics[scale=0.3]{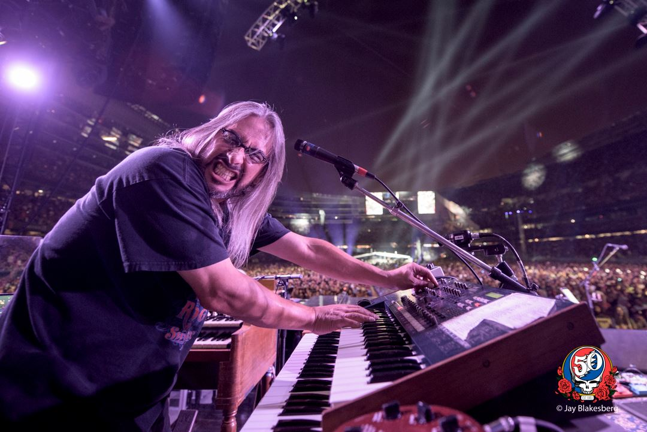} \\
User    & What is the position of the crowd in relation to the person on stage? \\
 & A. To the left B. To the right C. Behind D. In front\\
\midrule
InstructBLIP   & C. Behind \\
\midrule
Qwen-VL & D. In front \\
\midrule
LLaVA1.5 & D. In front \\
\midrule
MoAI & C. Behind \\
\bottomrule
\end{tabular}
}
\end{table}
\newpage
\begin{table}[h!]
\centering
\resizebox{\linewidth}{!}{
\renewcommand{\tabcolsep}{1.5mm}
\begin{tabular}{ll}
\toprule
\multicolumn{2}{l}{SEED Visual input example, Instance Interaction:} \\
\midrule
    & \includegraphics[scale=0.4]{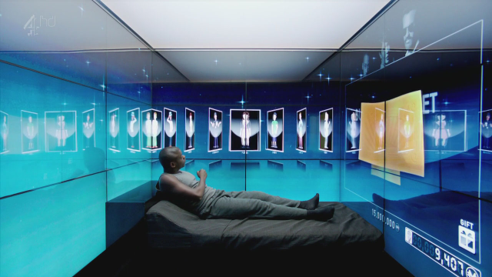} \\
User    & What is the person on the bed doing? \\
 & A. Sitting up B.Reading C. Watching TV D. Sleeping\\
\midrule
InstructBLIP   & D. Sleeping \\
\midrule
Qwen-VL & D. Sleeping \\
\midrule
LLaVA1.5 & D. Sleeping \\
\midrule
MoAI & C. Watching TV \\
\bottomrule
\end{tabular}
}
\end{table}
\begin{table}[h!]
\vspace{-5mm}
\centering
\resizebox{\linewidth}{!}{
\renewcommand{\tabcolsep}{1.5mm}
\begin{tabular}{ll}
\toprule
\multicolumn{2}{l}{SEED Visual input example, Text Recognition:} \\
\midrule
    & \includegraphics[scale=0.3]{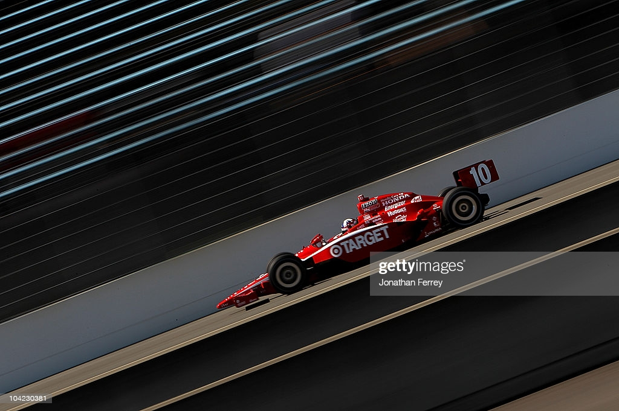} \\
User    & What color is the text on the sign in the image? \\
 & A. Black B. Red C. White D. Blue\\
\midrule
InstructBLIP   & B. Red \\
\midrule
Qwen-VL & B. Red \\
\midrule
LLaVA1.5 & B. Red \\
\midrule
MoAI & C. White \\
\bottomrule
\end{tabular}
}
\end{table}
\newpage
\begin{table}[h!]
\centering
\resizebox{\linewidth}{!}{
\renewcommand{\tabcolsep}{1.5mm}
\begin{tabular}{ll}
\toprule
\multicolumn{2}{l}{MM-Vet Visual input example, Recognition \& Spatial Awareness:} \\
\midrule
    & \includegraphics[scale=0.1]{figure/appendix9.jpg} \\
User    & On the right desk, what is to the left of the laptop? \\
\midrule
InstructBLIP   & Printer. \\
\midrule
Qwen-VL & Printer. \\
\midrule
LLaVA1.5 & There is a lamp to the left of the laptop on the right desk. \\
\midrule
MoAI & A Lamp. \\
\bottomrule
\end{tabular}
}
\end{table}
\newpage
\begin{table}[h!]
\centering
\resizebox{\linewidth}{!}{
\renewcommand{\tabcolsep}{3.5mm}
\begin{tabular}{l p{7cm}}
\toprule
\multicolumn{2}{l}{MM-Vet Visual input example, Recognition \& OCR:} \\
\midrule
    & \includegraphics[scale=0.25]{figure/appendix10.jpg} \\
User    & Can you explain this meme? \\
\midrule
InstructBLIP   & No. \\
\midrule
Qwen-VL & Spongebob and patrick write down what the professor says. \\
\midrule
LLaVA1.5 & The meme features a cartoon of two characters, one of which is a spongebob-like character, standing in front of a campfire. The other character is a professor, who is holding a clipboard and appears to be writing something. The scene is set in a camping environment, with a tent in the background.The meme is captioned with a humorous message, suggesting that the professor is writing a paper about the spongebob-like character. The text reads, Professor: ``Good morning, freshmen''. \\
\midrule
MoAI & The image is a meme featuring the cartoon characters Spongebob Squarepants and Patrick Star from the television show Spongebob Squarepants. Spongebob is holding a telescope and looking through it, while Patrick is standing next to him holding a piece of paper. The caption above the image reads Professor: ``Good morni...'' and the caption below the image reads Freshmen: Write that down, write that down!. The meme is a humorous take on the idea of students taking notes in class. \\
\bottomrule
\end{tabular}
}
\end{table}
\newpage
\begin{table}[h!]
\centering
\resizebox{\linewidth}{!}{
\renewcommand{\tabcolsep}{1.5mm}
\begin{tabular}{ll}
\toprule
\multicolumn{2}{l}{MM-Vet Visual input example, OCR \& Spatial Awareness:} \\
\midrule
    & \includegraphics[scale=0.15]{figure/appendix11.jpg} \\
User    & What is in the right most bottle? \\
\midrule
InstructBLIP   & Conditioner. \\
\midrule
Qwen-VL & Thicker \& fuller densifying shampoo. \\
\midrule
LLaVA1.5 & The right most bottle contains a tinted moisturizer \\
\midrule
MoAI & The right most bottle contains Thicker \& Fuller Mousse. \\
\bottomrule
\end{tabular}
}
\end{table}
\begin{table}[h!]
\vspace{-1mm}
\centering
\resizebox{\linewidth}{!}{
\renewcommand{\tabcolsep}{4mm}
\begin{tabular}{l p{7cm}}
\toprule
\multicolumn{2}{l}{MM-Vet Visual input example, OCR \& Math:} \\
\midrule
    & \includegraphics[scale=0.35]{figure/appendix12.png} \\
User    & What is the price difference between diesel and regular gasoline? \\
\midrule
InstructBLIP   & 0 \\
\midrule
Qwen-VL & 0.99 \\
\midrule
LLaVA1.5 & The price difference between diesel and regular gasoline is displayed on the gas pump. In the image, the diesel price is \$3.60, while the regular gasoline price is \$3.40. This means that diesel is slightly more expensive than regular gasoline. \\
\midrule
MoAI & 0.11 \\
\bottomrule
\end{tabular}
}
\end{table}
\end{document}